\documentclass[10pt,journal,compsoc]{IEEEtran}

\usepackage{cite}

\ifCLASSINFOpdf
  \usepackage[pdftex]{graphicx}
  \DeclareGraphicsExtensions{.pdf,.jpeg,.png}
\else
  \usepackage[dvips]{graphicx}
  \DeclareGraphicsExtensions{.eps}
\fi

\usepackage{amsmath}
\usepackage{amssymb}
\usepackage{amsfonts}

\usepackage{algorithmic}

\usepackage{textcomp}

\usepackage{url}

\usepackage[T1]{fontenc}
\usepackage[utf8]{inputenc}
\usepackage{xcolor}
\usepackage{booktabs}
\usepackage{subcaption}
\usepackage{multirow}
\usepackage{enumitem}
\usepackage{ragged2e}

\usepackage[colorlinks=true,linkcolor=blue,citecolor=blue,urlcolor=blue,breaklinks=true]{hyperref}

\usepackage{newunicodechar}
\newunicodechar{پ}{P}
\newunicodechar{چ}{Ch}
\newunicodechar{ژ}{Zh}
\newunicodechar{گ}{G}
\newunicodechar{ب}{B}
\newunicodechar{ت}{T}
\newunicodechar{ث}{S}
\newunicodechar{ل}{L}
\newunicodechar{ا}{A}
\newunicodechar{ـ}{-}
\newunicodechar{َ}{a}
\newunicodechar{ِ}{e}
\newunicodechar{ُ}{o}

\begin{document}

\title{Persian Pixel: A Large-Scale Synthetic OCR Dataset for Persian}

\author{Pouria~Mahdi*~and~Haq~Nawaz~Malik%
\IEEEcompsocitemizethanks{%
\IEEEcompsocthanksitem P.~Mahdi \textit{ORCID: 0009-0005-5472-1764
}.\protect\\
\IEEEcompsocthanksitem H.~N.~Malik \textit{ORCID: 0009-0003-1994-7640}.\protect\\
\IEEEcompsocthanksitem Hugging face Username: @{PouriaMahdi84} \& @{Omarrran}.\protect\\
\IEEEcompsocthanksitem The Persian Pixel dataset is publicly available at\protect\\
\protect\url{https://huggingface.co/datasets/Omarrran/Persian_Pixel}.}%
\thanks{}}

\IEEEtitleabstractindextext{
\begin{abstract}
\justifying
Optical Character Recognition (OCR) for Persian remains substantially less mature than for Latin-script languages despite Persian being spoken by more than 110 million people across multiple countries. This gap arises from two fundamental challenges: the intrinsic complexity of the Perso-Arabic writing system and the limited availability of large-scale, high-quality annotated datasets. Persian script exhibits obligatory cursive connectivity, context-dependent glyph shaping, extensive ligatures, diacritic placement, and stylistic variation across writing forms such as Naskh and Nastaliq, all of which significantly complicate text recognition. At the same time, the high cost and labor-intensive nature of manual annotation have created a persistent data bottleneck, limiting the development of robust OCR systems and slowing progress in Persian document digitization.In this paper, we introduce \textbf{Persian Pixel}, a comprehensive synthetic OCR dataset specifically designed to address these challenges. Comprising over 343,000 high-fidelity image text pairs, the dataset spans sentence, paragraph, and full-page document layouts generated from a carefully curated seven-million-word Persian corpus using the  SynthOCR-Gen rendering framework. The generation pipeline faithfully models the typographic characteristics of Persian script, including contextual character joining, positional glyph variants, diacritic placement, and multiple representative Persian typefaces. To bridge the synthetic-to-real domain gap, the rendered images are further enriched with more than twenty-five stochastic degradation models that emulate realistic document acquisition artifacts, including ink bleed, paper aging, blur, illumination variation, scanner imperfections, compression artifacts, and multiple noise processes.By overcoming the long-standing scarcity of annotated Persian OCR data, Persian Pixel provides a scalable and openly available resource for training and fine-tuning modern OCR architectures, including transformer-based models such as TrOCR and Donut. The dataset establishes a strong foundation for research in Persian document analysis, historical manuscript digitization, and end-to-end document understanding, while demonstrating that programmatic synthetic data generation offers a practical, cost-effective, and scalable alternative to manual annotation for advancing OCR in low-resource and typographically complex scripts.

\end{abstract}

}

\maketitle

\IEEEdisplaynontitleabstractindextext
\IEEEpeerreviewmaketitle

\section{Introduction}
The benefits of large-scale digitization have accrued unevenly across the world's languages. Scripts for which mature recognition pipelines and abundant training data exist enjoy near-frictionless conversion of print into searchable, machine-readable text, whereas languages whose writing systems are structurally more complex, or simply less resourced, remain partly excluded from that transformation. Persian is a telling case: spoken by more than 110 million people across Iran, Afghanistan, Tajikistan, and a large diaspora, and heir to one of the world's richest literary traditions, it is nonetheless served by Optical Character Recognition (OCR) systems that lag well behind their Latin-script counterparts. The practical consequence is that vast holdings of historical archives, government records, and contemporary print remain difficult to search, index, or preserve at scale.

The proximate cause of this gap is not a lack of algorithms but a lack of data. Modern recognition models are data-hungry, yet no large, high-quality, openly licensed corpus of annotated Persian document images has been available to train them. Manual annotation, the traditional remedy, is prohibitively slow and expensive for a script as visually intricate as Perso-Arabic, where a single base letter may surface as several distinct glyphs. The result is a self-reinforcing \textit{data deadlock}: scarce data yields brittle models that generalize poorly across fonts, degraded scans, and mixed-script content, and the resulting poor performance further discourages the investment needed to create better data. Existing Persian resources illustrate the point, being typically small, narrow in font coverage, or confined to a single document domain~\cite{persian_ocr_limitations, idplpfod}.

Synthetic data generation offers the most credible route out of this deadlock. By programmatically compositing text onto image canvases with exact ground truth, one obtains, at negligible marginal cost, perfectly labelled samples whose linguistic content, typography, and degradation can all be controlled. The strategy is well established for Latin scene text, where engines that render millions of synthetic words have driven substantial gains in recognition accuracy~\cite{jaderberg, synthtext}; for low-resource and typographically complex scripts, where real annotated data is essentially absent, it is often the only feasible path to a robust recognizer rather than merely a convenient one.

This paper introduces \textbf{Persian Pixel}, a large-scale synthetic dataset built to break the Persian OCR data deadlock. Our contributions are as follows.
\begin{itemize}[leftmargin=*]
    \item We release the first large-scale, multi-granularity synthetic dataset for Persian OCR: over 343{,}000 image--text pairs spanning sentence, paragraph, and page levels, each with pixel-accurate ground truth and structured metadata.
    \item We publicly release an openly licensed resource~\cite{persianpixel_data} ready for fine-tuning transformer recognizers such as TrOCR~\cite{trocr} and Donut~\cite{donut}, and we analyze its statistical and visual diversity to characterize its coverage and limits.
\end{itemize}

The remainder of the paper is organized as follows. Section~\ref{sec:background} reviews the evolution of OCR and situates our work within prior Persian and synthetic-data efforts. Section~\ref{sec:linguistic_analysis} analyzes the features of the Persian script that make recognition hard and that our generation pipeline must reproduce. Sections~\ref{sec:corpus}, \ref{sec:rendering}, and~\ref{sec:augmentation} describe, respectively, corpus acquisition, the SynthOCR-Gen rendering pipeline, and the visual augmentation suite. Section~\ref{sec:dataset_analysis} reports statistics and diversity metrics; Section~\ref{sec:discussion} discusses impact, use cases, and limitations; and Section~\ref{sec:conclusion} concludes and outlines future directions.

\section{Background and Related Work}
\label{sec:background}

\subsection{From Template Matching to Sequence Transducers}

OCR has passed through several distinct methodological eras. Early systems relied on template matching, hand-engineered features, and statistical classifiers, and were consequently fragile under variation in font, size, and degradation~\cite{early_ocr}. Hidden Markov Models introduced a probabilistic treatment of sequential text and, importantly, demonstrated segmentation-free recognition of connected scripts, including omnifont Arabic~\cite{hmm_ocr}. The decisive shift came with deep learning: convolutional feature extractors coupled with recurrent layers and a Connectionist Temporal Classification objective, the CRNN family, enabled fully end-to-end learning from pixels to character sequences without explicit segmentation, which is precisely the property that cursive scripts demand~\cite{crnn_ocr}. More recently, self-attention has supplanted recurrence: TrOCR couples a pre-trained vision transformer encoder with a text-transformer decoder and attains state-of-the-art results on printed and handwritten recognition~\cite{trocr}, while Donut dispenses with a separate OCR stage entirely, reading document images end-to-end for understanding tasks~\cite{donut}. These models are powerful, but their strength is largely inherited from pre-training on enormous Latin-script corpora, and that provenance is also their weakness: transferred to a right-to-left, obligatorily cursive script such as Perso-Arabic, accuracy degrades sharply unless substantial in-domain data is supplied.

\subsection{Persian OCR Resources and Their Limits}

Progress on Persian recognition has been constrained less by modelling than by data. Surveys of the field consistently note that available datasets are small, cover only a handful of fonts, and are often tied to a single document type, so that models trained on them overfit to particular typefaces and fail to generalize~\cite{persian_ocr_limitations, rahmati_persian}. Recent efforts have begun to enlarge the pool: the IDPL-PFOD line of work provides printed-Farsi text-line images at increasing scale~\cite{idplpfod, idplpfod2}, and community resources such as Arshasb contribute word-level annotated pages~\cite{arshasb}. These are valuable contributions, yet they remain predominantly Naskh-oriented, offer limited control over degradation, and rarely span multiple granularities within a single, openly licensed release. The situation parallels that of the closely related Urdu Nastaliq script, where a scarcity of annotated line images long held back recognition until purpose-built datasets and sequence models narrowed the gap~\cite{naz_urdu}. A large-scale, multi-font, multi-granularity, openly licensed Persian benchmark that deliberately includes Nastaliq and controlled degradation has, to our knowledge, been absent, and it is this gap that Persian Pixel is designed to fill.

\subsection{Synthetic Data for Text Recognition}

Where real annotation is impractical, synthetic generation has repeatedly proven to be an effective substitute. Jaderberg et al.\ showed that recognizers trained entirely on rendered word images could match models trained on real photographs~\cite{jaderberg}, and SynthText extended the idea to full scenes by compositing text into natural images with geometrically plausible placement~\cite{synthtext}. General-purpose tools such as TextRecognitionDataGenerator have since made synthetic corpora routine for many languages~\cite{trdg}. For complex cursive scripts, however, naive rendering is insufficient: correct output requires a shaping engine that resolves contextual forms, ligatures, and diacritic positioning from Unicode input. SynthOCR-Gen~\cite{synthocrgen} is built precisely for this regime, pairing a full OpenType shaping stack with a configurable degradation pipeline, which makes it well suited to generating faithful Perso-Arabic imagery at scale. The prerequisite for using it well is a clean, diverse text corpus, which explains the recent emphasis on large open Persian text collections such as \textit{naab}~\cite{naab}, as well as the analogous corpus-building efforts that have underpinned progress for other under-resourced language families~\cite{indicbert}.

\section{The Persian Script: Recognition Challenges}
\label{sec:linguistic_analysis}

Persian is written in an extended Arabic alphabet whose typographic behaviour departs sharply from the Latin conventions around which most recognition pipelines were designed. Understanding these behaviours is not incidental: they dictate both why generic OCR fails on Persian and what a faithful synthetic generator must reproduce. We group the salient challenges into four interacting phenomena: cursive glyph shaping, calligraphic style, sub-glyph disambiguation, and bidirectional layout.

\subsection{Cursivity, Contextual Forms, and Ligatures}

The Persian alphabet comprises thirty-two primary letters, four of which (پ, چ, ژ, گ) are absent from standard Arabic. Its defining property is obligatory cursivity: within a word, joining letters connect to their neighbours, and each base character therefore assumes up to four position-dependent shapes: isolated, initial, medial, and final. The letter \textit{Beh} (ب, U+0628), for instance, is realized differently in isolation, at a word's start, in its interior, and at its end. Across the alphabet this expands roughly thirty-two base characters into well over a hundred visual glyphs, so that the mapping from Unicode codepoints to rendered forms is many-valued and context-sensitive. Cursivity also mandates certain ligatures, most prominently \textit{Lam-Aleph} (لا), in which the sequence \textit{Lam} (U+0644) followed by \textit{Aleph} (U+0627) must be both rendered and recognized as a single indivisible unit. For a recognizer, the immediate implication is that character segmentation in the Latin sense is ill-defined; end-to-end sequence models are effective on Persian precisely because they sidestep it, and a synthetic generator that composited letters independently would produce disjointed, unusable imagery.

\subsection{The Naskh--Nastaliq Divide}

Two calligraphic traditions dominate Persian typesetting, and they impose very different demands. \textbf{Naskh} is upright and comparatively geometric, sits on a stable horizontal baseline, and prevails in modern digital typography, newspapers, and printed books; its regularity makes it the easier target for automatic recognition. \textbf{Nastaliq}, by contrast, is the canonical style of classical poetry and much historical material and is visually far more demanding: it hangs words from a sloping, right-to-left descending baseline, stacks letter components vertically, and allows strokes to overlap between adjacent glyphs. Models trained on Naskh alone therefore tend to collapse on Nastaliq, whose two-dimensional layout violates the near-linear arrangement they have learned to expect. The same effect is documented in the closely related Urdu Nastaliq setting, where dedicated architectures and data were required to make progress~\cite{naz_urdu}. Deliberately representing both styles is thus essential to any dataset that aims at general Persian coverage.

\subsection{Dots, Diacritics, and the Noise Problem}

A distinctive fragility of Perso-Arabic recognition arises from its heavy reliance on small marks. Several letters are distinguished only by the number and placement of dots: \textit{Beh} (ب), \textit{Peh} (پ), \textit{Teh} (ت), and \textit{Seh} (ث) share an identical skeleton and differ solely in their dot pattern, so that a single lost or merged dot silently converts one letter into another. Because these marks are small and high in spatial frequency, they are exactly what document degradation attacks first: blurring, ink spread, and sensor noise routinely erase or confuse them, and the resulting substitutions propagate into cascading word-level errors. Optional short-vowel diacritics (\textit{harakat}), namely \textit{fathah} (َ, U+064E), \textit{kasrah} (ِ, U+0650), and \textit{dammah} (ُ, U+064F), compound the difficulty: usually omitted in everyday text but obligatory in religious, classical, and pedagogical material, they alter both appearance and reading, so a recognizer must handle their presence and absence gracefully. This tight coupling between fine detail and semantic identity is a principal reason our augmentation suite (Section~\ref{sec:augmentation}) models dot- and stroke-level degradation explicitly.

\subsection{Right-to-Left and Bidirectional Layout}

Persian is written right-to-left, but contemporary documents freely embed left-to-right runs, such as Latin words, Arabic and Latin numerals, URLs, and mathematical expressions, producing genuinely bidirectional text. In such content the logical order of characters in memory differs from their visual order on the page, and correct rendering requires the Unicode Bidirectional Algorithm to resolve the two~\cite{bidi}. Pipelines built for unidirectional Latin text mishandle these transitions, emitting scrambled or misordered output at script boundaries. Faithful synthesis, and by extension robust recognition, therefore depends on applying the bidirectional algorithm during layout so that ground-truth strings and rendered images agree in both logical content and visual arrangement.

\section{Methodology I: Corpus Acquisition and Curation}
\label{sec:corpus}

The linguistic quality of a synthetic OCR dataset is bounded by the quality of the text it renders: whatever vocabulary, register, and orthographic variety the corpus lacks, the trained recognizer will never see. For Persian Pixel we therefore assembled a corpus of more than seven million words, drawn exclusively from open-licensed web sources to permit unrestricted redistribution, and curated it to reflect the breadth of contemporary and historical Persian rather than any single register.

Domain diversity was a primary objective, since a corpus skewed toward one register yields models that falter on others. The collection spans encyclopedic prose (Persian Wikipedia), providing broad factual and scientific vocabulary; literary material, from classical poetry to modern fiction, contributing archaic forms and varied syntax; news media, reflecting current journalistic usage; government and administrative portals, supplying formal terminology; curated public social-media text, capturing informal and colloquial language; and digitized historical documents, which introduce older orthography and specialized lexis. This mixture parallels the composition of recent large open Persian corpora~\cite{naab} and is intended to expose the recognizer to the full stylistic range it will encounter in real documents.

Raw web text is unusable without substantial cleaning, so we applied a multi-stage pipeline before rendering. Corrupted encodings (\textit{mojibake}) were detected and repaired to guarantee valid UTF-8, and residual HTML, XML, and other markup was stripped to recover plain text. Personally identifiable information, such as names, e-mail addresses, and telephone numbers, was detected and removed to protect privacy. A script-purity filter discarded segments dominated by non-Perso-Arabic Unicode blocks, retaining text composed chiefly of Persian letters, numerals, and punctuation. All retained text was then normalized to Unicode Normalization Form C, which is essential for consistent shaping and for exact string comparison in the presence of combining diacritics, and whitespace was standardized to collapse redundant spacing. Finally, exact and near-duplicate sentences and paragraphs were removed to curb redundancy and mitigate memorization, and segments that were too short to be linguistically coherent or too long for stable rendering were filtered out.

The curated corpus retains over seven million words with a large unique vocabulary and a balanced distribution across the domains above. Because the visual dataset inherits its linguistic characteristics directly from this text, corpus cleanliness and diversity are not preliminaries to the work but determinants of the recognizer's eventual ability to generalize across the Persian written landscape.

\section{Methodology II: The SynthOCR-Gen Rendering Pipeline}
\label{sec:rendering}

Persian Pixel is rendered with \textit{SynthOCR-Gen}~\cite{synthocrgen}, an engine engineered to reproduce the typographic behaviour of low-resource cursive scripts rather than merely to place characters on a canvas. As established in Section~\ref{sec:linguistic_analysis}, the correctness of Perso-Arabic imagery depends on contextual shaping, ligature formation, diacritic positioning, and bidirectional layout; SynthOCR-Gen addresses these by coupling a full text-shaping stack to a configurable rendering front end, which is what allows the synthetic-to-real domain gap to be kept small.

\subsection{Shaping-Aware Rendering}

The fidelity of the output rests on SynthOCR-Gen's shaping-aware rendering, which resolves a Unicode string into correctly formed and positioned glyphs before rasterization rather than compositing characters independently. This shaping step is what makes the imagery usable for training: it selects each letter's isolated, initial, medial, or final form from context, substitutes multi-character sequences such as \textit{Lam-Aleph} with their single ligature glyph, places \textit{harakat} at their proper offsets, and realizes the sloping baseline and vertical stacking of Nastaliq. Crucially, it also applies the Unicode Bidirectional Algorithm so that mixed right-to-left and left-to-right runs are ordered consistently between the rendered image and its ground-truth string. Without this shaping stage, a naive concatenation of per-character bitmaps would yield disconnected, misshapen text bearing no relation to real Persian print; such imagery would actively mislead a recognizer rather than train it. It is precisely because SynthOCR-Gen encapsulates this behaviour for low-resource cursive scripts~\cite{synthocrgen} that it, rather than a generic text-rendering tool, is used to generate Persian Pixel.

\subsection{Typographic Coverage: A Seven-Font Strategy}

Because recognizers trained on few typefaces overfit to them, we render every corpus segment across seven fonts chosen to span the stylistic range of real Persian documents. The set comprises three modern Naskh faces optimized for print and screen (\textit{Vazirmatn}, \textit{Sahel}, and \textit{Shabnam}), the traditionally styled Naskh face \textit{Amiri}, the dedicated Nastaliq face \textit{IranNastaliq}, which contributes the diagonal baseline and vertical stacking absent from the others, and two further display and print faces that broaden the typographic spectrum. Exposing the same words to these varied realizations pushes a model toward abstract, font-invariant features rather than memorized templates, directly countering the narrow font coverage that limits earlier Persian resources (Section~\ref{sec:background}).

\subsection{Multi-Granularity Rendering}

To support the full spectrum of recognition tasks, Persian Pixel provides images at three granularities. \textbf{Sentence-level} samples (251{,}000 rows, typical width 640\,px) contain a single line or short sentence and target line-level recognizers, the workhorse of most OCR pipelines. \textbf{Paragraph-level} samples (110{,}138 rows, typical width 512\,px) span several lines, exercising line-break handling and multi-line context. \textbf{Page-level} samples (31{,}108 rows, typical width 768\,px) emulate whole documents and suit end-to-end models such as Donut~\cite{donut} together with layout-analysis and multi-column reading tasks. This layering lets practitioners match training data to the granularity of their target application, from precise line transcription to holistic document understanding.

\subsection{Annotation Schema}

Every image ships with pixel-accurate ground truth and structured metadata sufficient for supervised training and downstream analysis. Each record carries the raw image \texttt{bytes} and its \texttt{path}; the exact rendered Unicode \texttt{text} that serves as the recognition target; a \texttt{sample\_type} tag recording granularity (\texttt{sentence}, \texttt{paragraph}, or \texttt{page}); a \texttt{source\_run\_id} linking the sample to its generation configuration for reproducibility; the image \texttt{width} and \texttt{height} in pixels; and the derived \texttt{text\_chars} and \texttt{line\_count} statistics. Together these fields provide everything required to train, filter, and stratify recognizers on the dataset.

\section{Methodology III: The Visual Augmentation Suite}
\label{sec:augmentation}

Accurate shaping guarantees typographically correct text, but real documents are rarely pristine. Scanning, printing, handling, and aging introduce distortions that a recognizer trained only on clean renders will never have learned to tolerate. To close this \textit{synthetic-to-real gap}, we subject every rendered image to a suite of more than twenty-five degradation operators, applied stochastically and in combination so that the effective sample space is combinatorially large and no two images degrade identically. The operators fall into four families, each targeting a different source of real-world variation.

\textbf{Geometric transformations} reproduce the spatial imperfections of capture and handling. Small random rotations (typically within $\pm5^{\circ}$) model scanner and camera misalignment; horizontal and vertical shear mimics imperfect paper feeding and oblique viewpoints; perspective warping via homographies simulates documents photographed at an angle or physically curved; and mild scaling and aspect-ratio jitter accounts for varying print sizes and optical distortion.

\textbf{Photometric distortions} model the appearance of a page under different lighting and imaging conditions. Brightness and contrast are randomly adjusted to span illumination levels and paper quality; color jitter perturbs hue, saturation, and value to emulate different inks, paper tints, and camera color profiles; and occasional inversion reproduces negative or specialized scanning modes.

\textbf{Noise injection} emulates sensor and print artifacts across their characteristic statistics. Additive Gaussian noise represents ordinary digital sensor noise; salt-and-pepper noise introduces sparse extreme-valued pixels standing in for dust, scratches, and sensor defects; Poisson noise captures the photon-limited regime of dim or aged scans; and multiplicative speckle adds granular texture. Motion blur, Gaussian blur, and median filtering complete the family, reproducing camera shake, defocus, and general loss of sharpness, the conditions under which the small dots and diacritics discussed in Section~\ref{sec:linguistic_analysis} are most likely to be lost.

\textbf{Document-specific degradations} imitate the physical wear of printed and historical material. Ink bleed, realized by morphological dilation, spreads strokes slightly beyond their outlines as on absorbent or aged paper; erosion and dilation more generally vary stroke width to mimic differing ink density and print quality. Rendering onto yellowed, stained, or textured paper backgrounds supplies realistic document context, while scan-line artifacts inject the periodic banding of imperfect scanners. Random erasing removes rectangular regions to simulate smudges, tears, and occlusions, and synthetic shadow and highlight fields reproduce uneven lighting and page curvature.

Considered together, the four families cover complementary axes of real-world variation (spatial, photometric, statistical, and physical), and their stochastic composition yields an immense diversity of degraded samples from a fixed corpus and font set. This is the mechanism by which Persian Pixel confronts a recognizer with the conditions of genuine scanned and photographed documents, and thereby the mechanism through which robustness and generalization are expected to emerge, an approach whose efficacy is well established for Latin scene text~\cite{jaderberg, synthtext}.

\section{Dataset Statistics and Diversity}
\label{sec:dataset_analysis}

Persian Pixel is characterized as much by its diversity as by its scale. This section quantifies both, reporting the distribution of samples across granularities and the lexical, script-level, and typographic variety that the generation pipeline produces.

\subsection{Granularity Distribution}

The dataset contains 343{,}246 unique image--text pairs partitioned across the three granularity levels introduced in Section~\ref{sec:rendering}, as summarized in Table~\ref{tab:granularity_distribution}. Sentence-level images dominate, reflecting their role as the primary training signal for line recognizers, while paragraph- and page-level images provide progressively richer layout context. The per-level counts sum to more than the number of unique samples because paragraph and page configurations are composed from a superset of the sentence-level text, so their underlying content overlaps; the percentages in Table~\ref{tab:granularity_distribution} are therefore expressed relative to the total and are not mutually exclusive.

\begin{table}[!t]
\caption{Persian Pixel Granularity Distribution}
\label{tab:granularity_distribution}
\centering
\begin{tabular}{lcc}
\toprule
\textbf{Sample Type} & \textbf{Rows} & \textbf{Share of Total} \\
\midrule
Sentence level       & 251,000                 & 73.1\%             \\
Paragraph level      & 110,138                 & 32.1\%             \\
Page level           & 31,108                  & 9.1\%              \\
\midrule
\textbf{Total unique samples} & \textbf{343,246}        & \textbf{100\%}             \\
\bottomrule
\end{tabular}
\end{table}

\subsection{Lexical Diversity}

Because the imagery inherits its content from the underlying seven-million-word corpus (Section~\ref{sec:corpus}), the dataset's linguistic richness is best read from that text. The corpus exhibits a high type--token ratio, indicating that its vocabulary is broad rather than a small set of words repeated at length, and it contains a substantial tail of hapax legomena (words occurring exactly once), which signals genuine lexical depth, including rare terms and proper nouns. Exposure to this long tail is what allows a trained recognizer to generalize beyond high-frequency vocabulary rather than overfitting to it.

\subsection{Script Purity}

The script-purity filter of Section~\ref{sec:corpus} is reflected in the composition of the rendered text: the overwhelming majority of characters lie within the Perso-Arabic Unicode blocks, with non-Perso-Arabic characters confined to the small residue of embedded foreign words, URLs, and technical terms that occur naturally in mixed-script documents. Whitespace and numerals, in both Eastern-Arabic and Latin forms, are well represented, matching their ordinary frequency in Persian text. The dataset thus concentrates the learning signal on Persian glyphs and their interactions while preserving the realistic minority of left-to-right content that a robust recognizer must also handle.

\subsection{Typographic and Visual Diversity}

Two further sources of variation distinguish Persian Pixel from prior Persian resources. Typographically, the seven-font strategy of Section~\ref{sec:rendering} spans modern Naskh (Vazirmatn, Sahel, Shabnam), classical Naskh (Amiri), and Nastaliq (IranNastaliq); the deliberate inclusion of Nastaliq, with its sloping baseline and vertical stacking, is particularly consequential, since that style is largely absent from existing datasets yet indispensable for classical and historical material. Visually, the stochastic composition of the twenty-five-plus augmentation operators (Section~\ref{sec:augmentation}) across seven fonts and three granularities yields an effectively unbounded space of distinct degraded images: one sentence may appear in Vazirmatn with ink bleed over aged paper, another in IranNastaliq under Gaussian noise and perspective warp. Because near-identical samples are essentially never produced, the dataset forces recognizers toward features invariant to typeface and degradation, spanning the continuum from pristine digital print to heavily degraded historical scans.

\section{Discussion: Impact, Use Cases, and Limitations}
\label{sec:discussion}

Beyond its technical specification, Persian Pixel carries socio-technical significance and enables a range of downstream applications, even as it leaves clear problems open. We consider each in turn.

\subsection{Broadening Access to Persian Language Technology}

Progress in NLP and document intelligence for low-resource languages has historically been gated by the availability of open, high-quality data, a bottleneck that concentrates capability in well-resourced institutions and encourages reliance on proprietary systems. By releasing a large, openly licensed dataset, Persian Pixel lowers that barrier for researchers, developers, and startups across Iran, Afghanistan, and Tajikistan, as well as the global diaspora, allowing them to build and adapt state-of-the-art recognizers without first solving the data problem themselves. The immediate benefits are practical, including searchable archives, digitized public records, and preserved cultural heritage, but the broader effect is to distribute the means of building Persian language technology more widely, in the spirit of the open-corpus efforts that have advanced other under-resourced language families~\cite{naab, indicbert}.

\subsection{Primary Use Cases}

The dataset is designed to support several concrete workflows. Its scale and diversity make it well suited to \textit{fine-tuning transformer recognizers}: models such as TrOCR~\cite{trocr} and Donut~\cite{donut}, pre-trained on Latin script or any other arabic native pretrainning model, can be adapted to Persian, including degraded and layout-rich documents, far more effectively with in-domain data of this size. Because degradation is controllable, the dataset naturally supports \textit{curriculum learning}, in which training begins on clean renders and advances to heavily augmented ones, a schedule that can stabilize optimization. It also serves as a \textit{pre-training substrate for domain adaptation}: a recognizer pre-trained on Persian Pixel and then fine-tuned on a small amount of real, domain-specific data can reach high accuracy while sharply reducing manual annotation. Finally, its exact ground truth makes it a convenient \textit{benchmark for post-OCR correction}: passing images through weaker OCR engines yields aligned pairs of noisy predictions and gold text, precisely the supervision needed to train and evaluate correction models.

\subsection{The Compute Barrier}

Solving the data problem exposes a second constraint. Fine-tuning large recognizers on hundreds of thousands of images demands substantial GPU resources and training time, which can place the work out of reach for individual researchers and smaller institutions. Several partial remedies exist and deserve emphasis: continued research into parameter-efficient and computationally lighter architectures suited to constrained settings; managed platforms that offer GPU access and pre-trained models as a service, lowering the barrier to experimentation and deployment; and public--private partnerships that pool compute and expertise for language-technology projects of public value. None fully dissolves the barrier, but together they widen access to the models that Persian Pixel is meant to enable.

\subsection{Limitations}

Three limitations bound the dataset's applicability and motivate future work. First, it is confined to \textit{printed and digitally rendered text}; handwriting recognition, a substantially harder problem for Persian, lies outside its scope. Second, its \textit{font coverage}, though broad by the standards of prior work, is constrained by the availability of high-quality, openly licensed Persian typefaces, so the long tail of stylized calligraphic and historical fonts is only partially represented. Third, its \textit{layout modelling} captures paragraph and page structure but not the intricate arrangements of many historical documents, such as marginalia, interlinear glosses, and non-linear reading orders, which a more advanced layout-generation stage would be needed to reproduce. These boundaries mark where further dataset and pipeline development is required to approach truly universal Persian document digitization.

\section{Conclusion}
\label{sec:conclusion}

We have presented Persian Pixel, a large-scale synthetic dataset built to break the data deadlock that has held back Persian OCR. Combining the shaping-aware SynthOCR-Gen engine, a curated seven-million-word corpus, a seven-font strategy that deliberately spans Naskh and Nastaliq, and a suite of more than twenty-five stochastic degradation operators, the dataset delivers over 343{,}000 image--text pairs whose linguistic content and visual variation reflect the genuine complexity of the Perso-Arabic script. In providing an openly licensed, multi-granularity resource ready for fine-tuning modern recognizers such as TrOCR~\cite{trocr} and Donut~\cite{donut}, it offers a practical foundation for advancing Persian document digitization and for broadening participation in Persian language technology.

Three directions extend this work. The first is \textit{generative handwriting synthesis}: diffusion-based methods have recently produced convincing styled handwritten text for Latin script~\cite{wordstylist}, and adapting them to Persian would carry the synthetic-data strategy into the far harder handwritten regime. The second is \textit{layout-aware modelling for historical archives}, developing recognizers that preserve the complex, often non-linear layouts of historical Persian documents rather than only transcribing their text. The third is \textit{cross-lingual transfer}, exploiting the shared Perso-Arabic script to carry models trained on Persian Pixel toward related low-resource languages such as Dari, Pashto, and Urdu, where the parallels in cursivity and Nastaliq styling are already well documented~\cite{naz_urdu}.

\section*{Acknowledgments}

The authors thank the developers of the SynthOCR-Gen rendering engine, on which the generation of this dataset relies, and the font designers who release their typefaces under open licenses, without whose generosity the typographic diversity of Persian Pixel would not be possible.

\section*{References}
\bibliographystyle{IEEEtran}
\bibliography{sections/references}

\appendices
\section{Representative Visual Samples}

\begin{figure*}[htbp]
\centering
\begin{subfigure}[b]{0.24\textwidth}\centering\includegraphics[width=\textwidth]{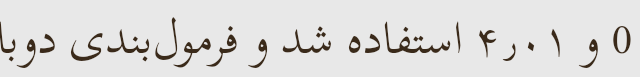}\caption{Sentence Sample 1}\end{subfigure}\hfill
\begin{subfigure}[b]{0.24\textwidth}\centering\includegraphics[width=\textwidth]{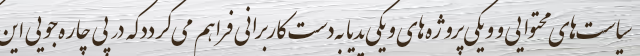}\caption{Sentence Sample 2}\end{subfigure}\hfill
\begin{subfigure}[b]{0.24\textwidth}\centering\includegraphics[width=\textwidth]{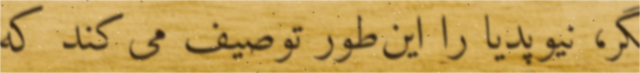}\caption{Sentence Sample 3}\end{subfigure}\hfill
\begin{subfigure}[b]{0.24\textwidth}\centering\includegraphics[width=\textwidth]{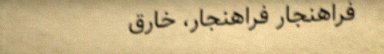}\caption{Sentence Sample 4}\end{subfigure}

\vspace{1em}
\begin{subfigure}[b]{0.24\textwidth}\centering\includegraphics[width=\textwidth]{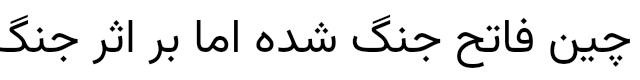}\caption{Sentence Sample 5}\end{subfigure}\hfill
\begin{subfigure}[b]{0.24\textwidth}\centering\includegraphics[width=\textwidth]{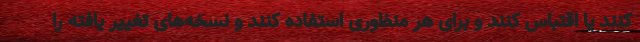}\caption{Sentence Sample 6}\end{subfigure}\hfill
\begin{subfigure}[b]{0.24\textwidth}\centering\includegraphics[width=\textwidth]{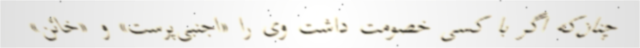}\caption{Sentence Sample 7}\end{subfigure}\hfill
\begin{subfigure}[b]{0.24\textwidth}\centering\includegraphics[width=\textwidth]{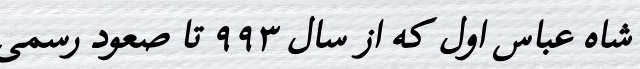}\caption{Sentence Sample 8}\end{subfigure}

\vspace{1em}
\begin{subfigure}[b]{0.24\textwidth}\centering\includegraphics[width=\textwidth]{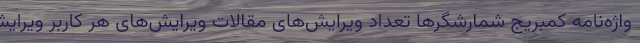}\caption{Sentence Sample 9}\end{subfigure}\hfill
\begin{subfigure}[b]{0.24\textwidth}\centering\includegraphics[width=\textwidth]{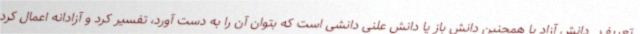}\caption{Sentence Sample 10}\end{subfigure}\hfill
\hspace{0.24\textwidth}\hfill
\hspace{0.24\textwidth}
\caption{Representative sentence-level samples from Persian Pixel, demonstrating varied fonts and subtle augmentations applied to single lines of text.}
\label{fig:sentence_samples}
\end{figure*}

\begin{figure*}[htbp]
\centering
\begin{subfigure}[b]{0.24\textwidth}\centering\includegraphics[width=\textwidth]{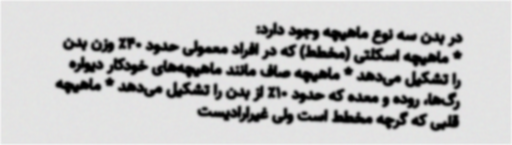}\caption{Paragraph Sample 1}\end{subfigure}\hfill
\begin{subfigure}[b]{0.24\textwidth}\centering\includegraphics[width=\textwidth]{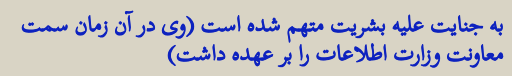}\caption{Paragraph Sample 2}\end{subfigure}\hfill
\begin{subfigure}[b]{0.24\textwidth}\centering\includegraphics[width=\textwidth]{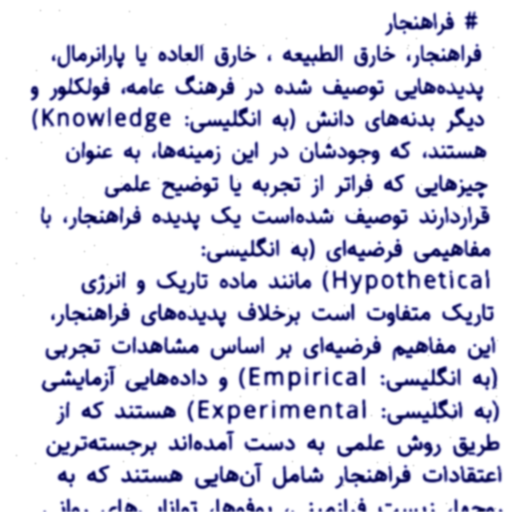}\caption{Paragraph Sample 3}\end{subfigure}\hfill
\begin{subfigure}[b]{0.24\textwidth}\centering\includegraphics[width=\textwidth]{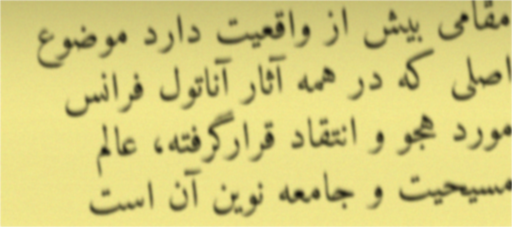}\caption{Paragraph Sample 4}\end{subfigure}

\vspace{1em}
\begin{subfigure}[b]{0.24\textwidth}\centering\includegraphics[width=\textwidth]{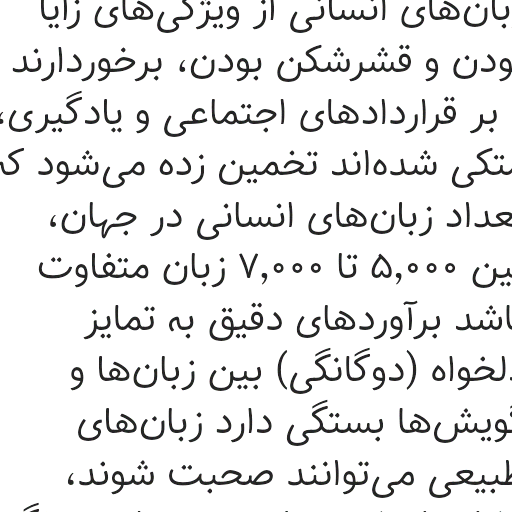}\caption{Paragraph Sample 5}\end{subfigure}\hfill
\begin{subfigure}[b]{0.24\textwidth}\centering\includegraphics[width=\textwidth]{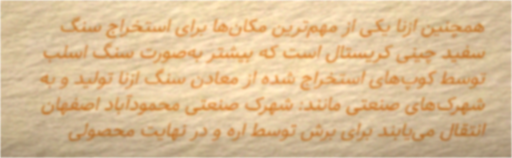}\caption{Paragraph Sample 6}\end{subfigure}\hfill
\begin{subfigure}[b]{0.24\textwidth}\centering\includegraphics[width=\textwidth]{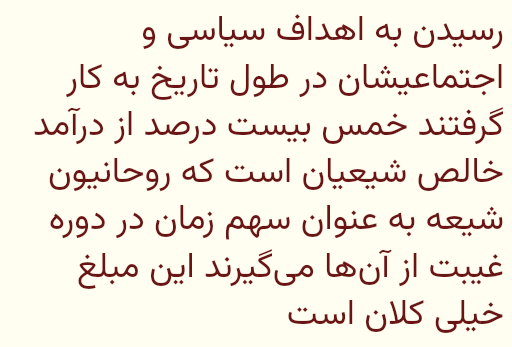}\caption{Paragraph Sample 7}\end{subfigure}\hfill
\begin{subfigure}[b]{0.24\textwidth}\centering\includegraphics[width=\textwidth]{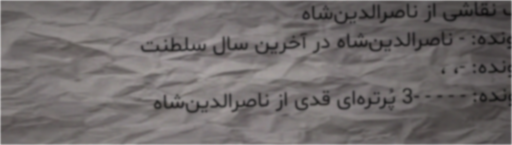}\caption{Paragraph Sample 8}\end{subfigure}

\vspace{1em}
\begin{subfigure}[b]{0.24\textwidth}\centering\includegraphics[width=\textwidth]{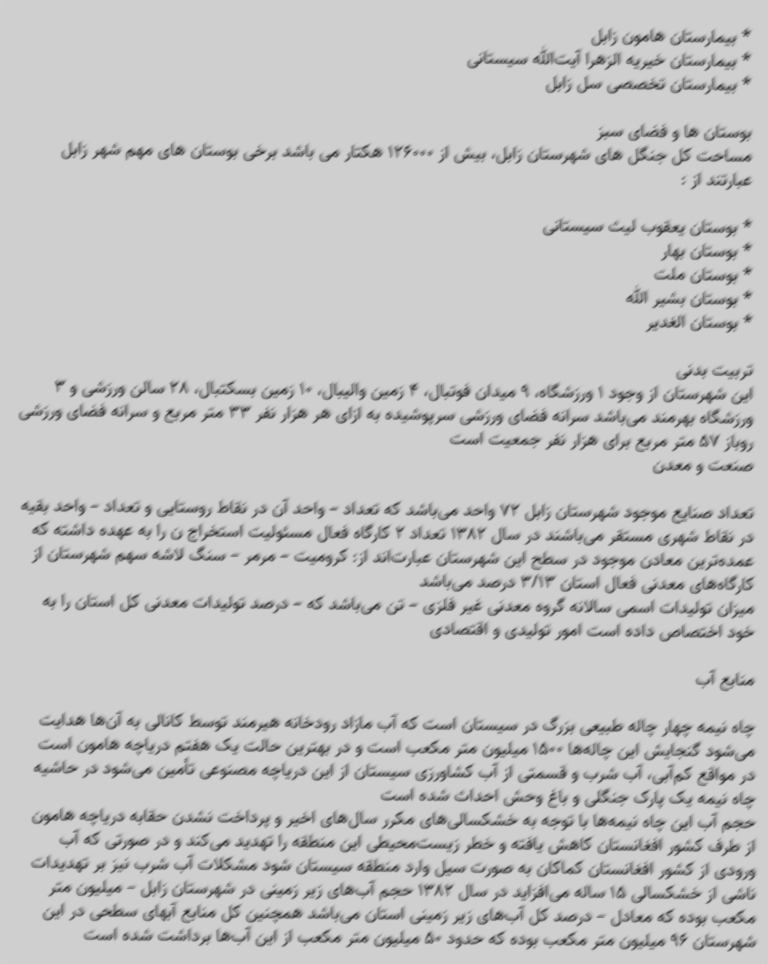}\caption{Paragraph Sample 9}\end{subfigure}\hfill
\begin{subfigure}[b]{0.24\textwidth}\centering\includegraphics[width=\textwidth]{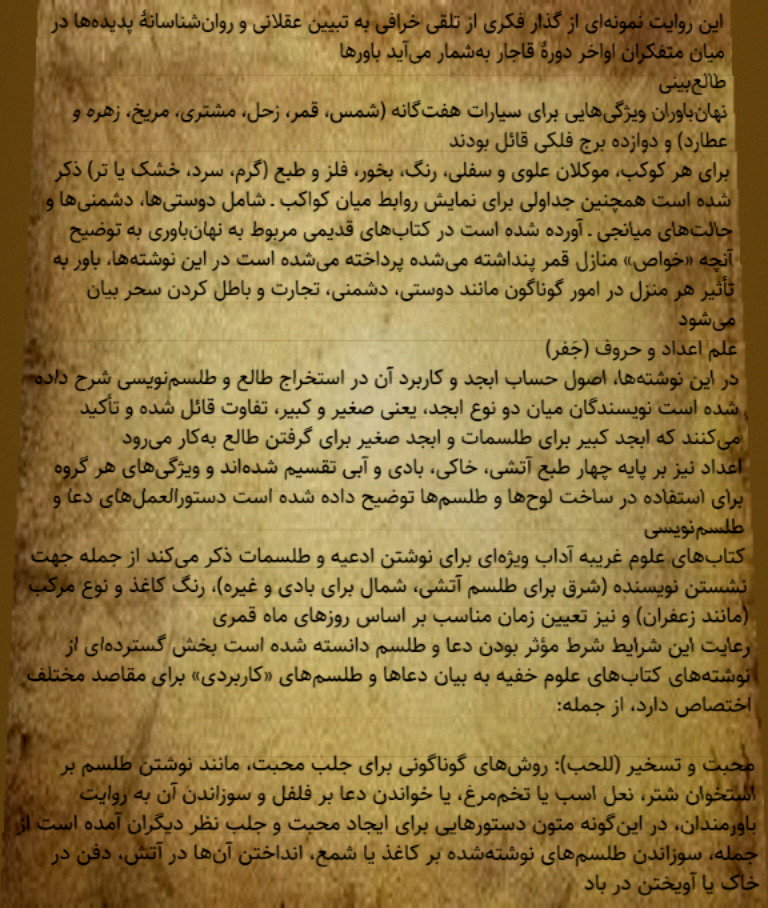}\caption{Paragraph Sample 10}\end{subfigure}\hfill
\begin{subfigure}[b]{0.24\textwidth}\centering\includegraphics[width=\textwidth]{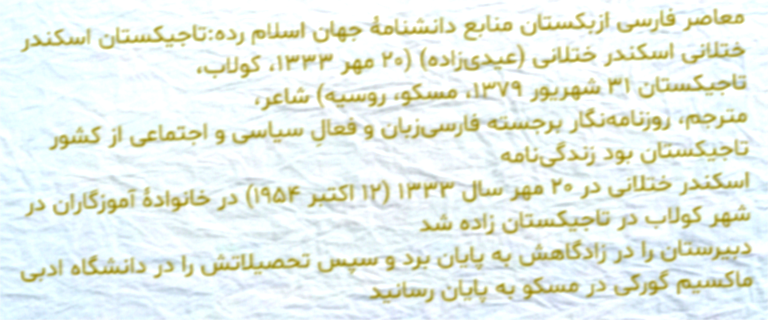}\caption{Paragraph Sample 11}\end{subfigure}\hfill
\hspace{0.24\textwidth}
\caption{Representative paragraph-level samples from Persian Pixel, illustrating multi-line text with varying layouts and degradation effects.}
\label{fig:paragraph_samples}
\end{figure*}

\begin{figure*}[htbp]
\centering
\begin{subfigure}[b]{0.30\textwidth}\centering\includegraphics[width=\textwidth]{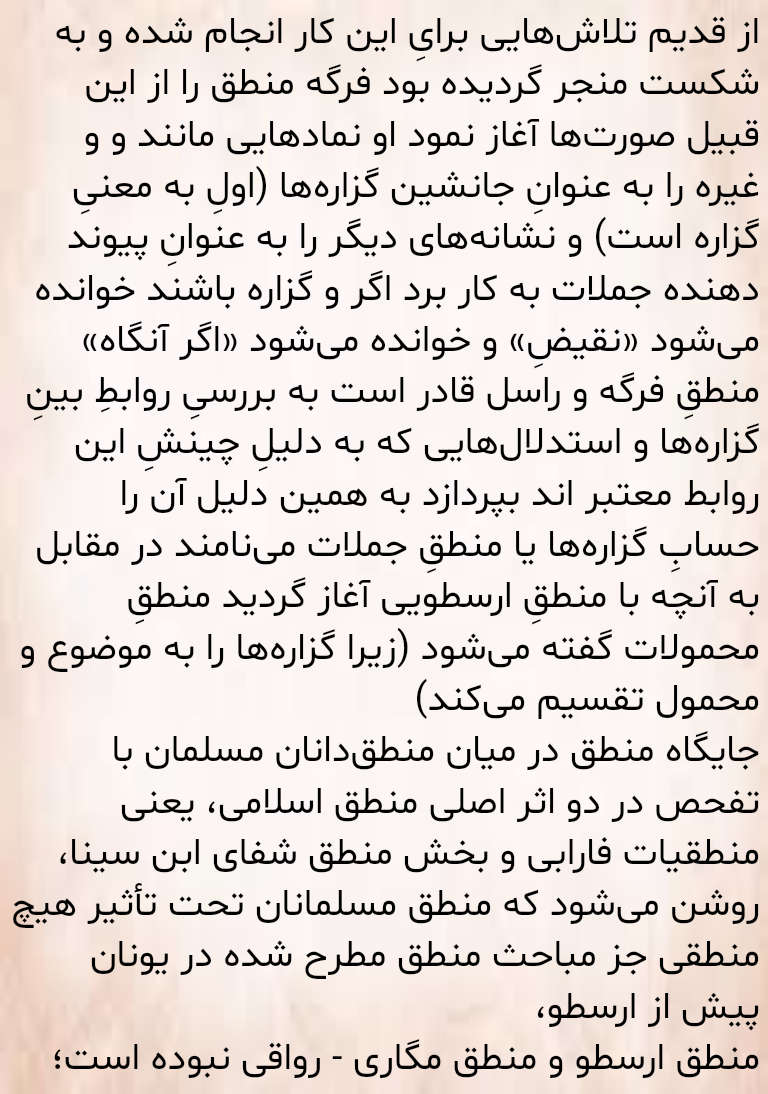}\caption{Page Sample 1}\end{subfigure}\hfill
\begin{subfigure}[b]{0.30\textwidth}\centering\includegraphics[width=\textwidth]{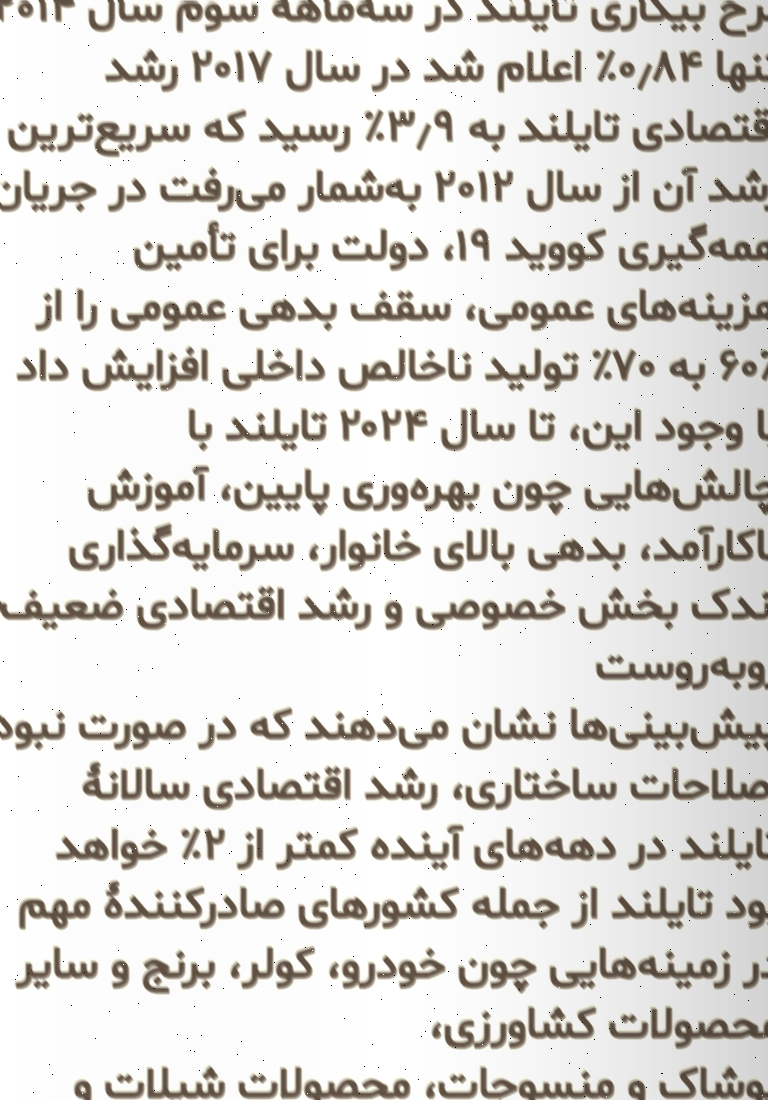}\caption{Page Sample 2}\end{subfigure}\hfill
\begin{subfigure}[b]{0.30\textwidth}\centering\includegraphics[width=\textwidth]{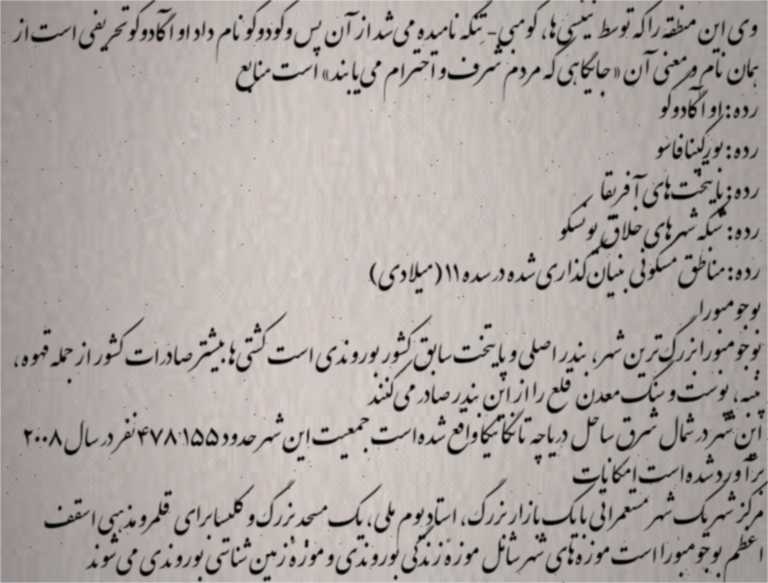}\caption{Page Sample 3}\end{subfigure}

\vspace{1em}
\begin{subfigure}[b]{0.30\textwidth}\centering\includegraphics[width=\textwidth]{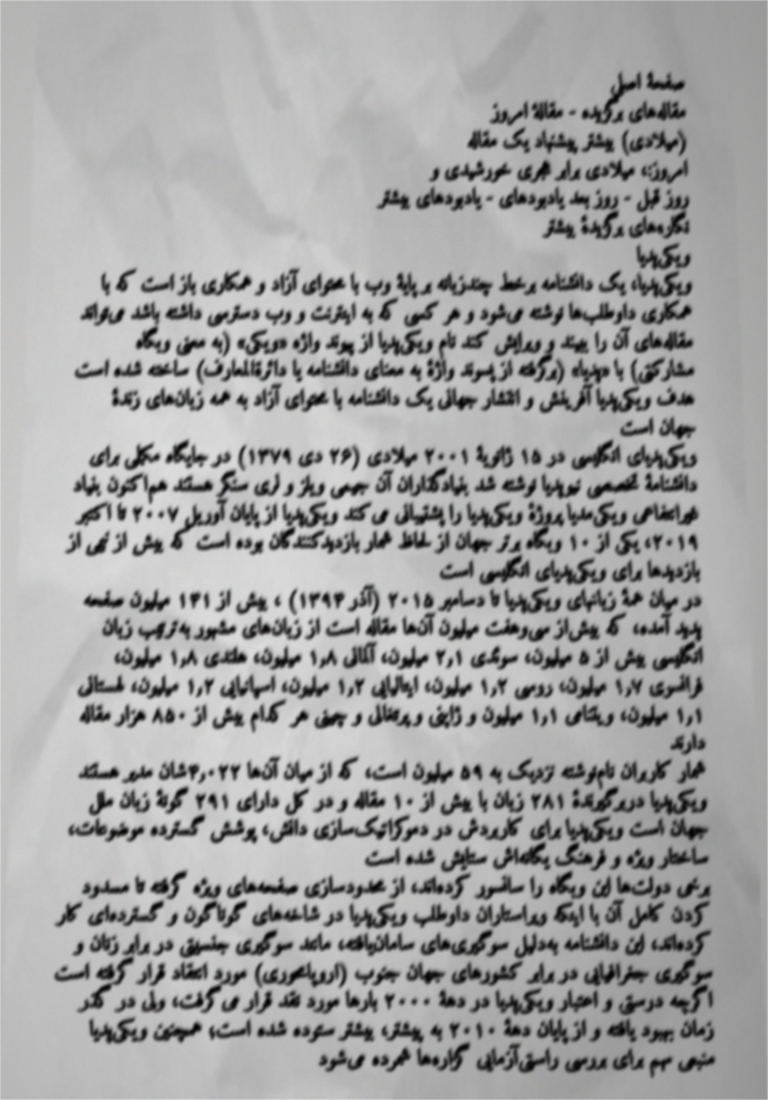}\caption{Page Sample 4}\end{subfigure}\hfill
\begin{subfigure}[b]{0.30\textwidth}\centering\includegraphics[width=\textwidth]{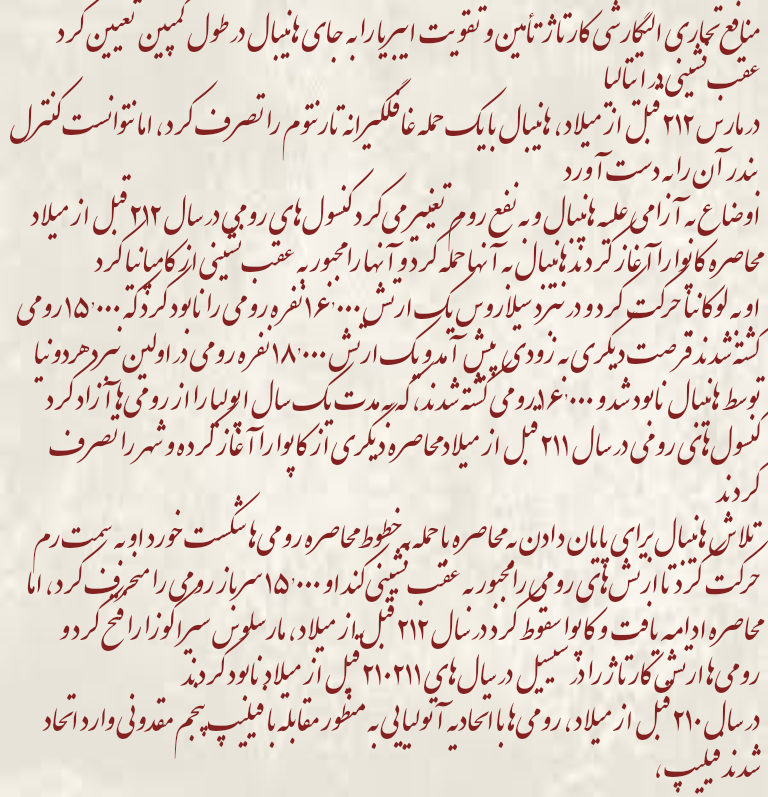}\caption{Page Sample 5}\end{subfigure}\hfill
\begin{subfigure}[b]{0.30\textwidth}\centering\includegraphics[width=\textwidth]{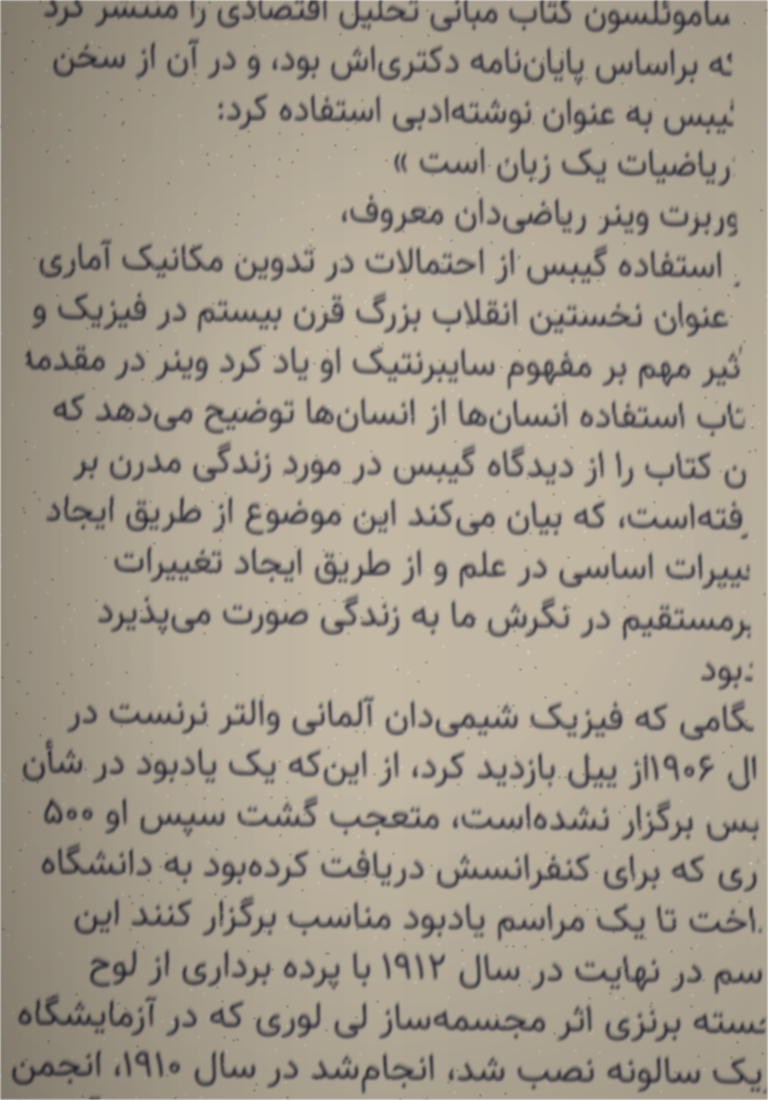}\caption{Page Sample 6}\end{subfigure}

\vspace{1em}
\begin{subfigure}[b]{0.30\textwidth}\centering\includegraphics[width=\textwidth]{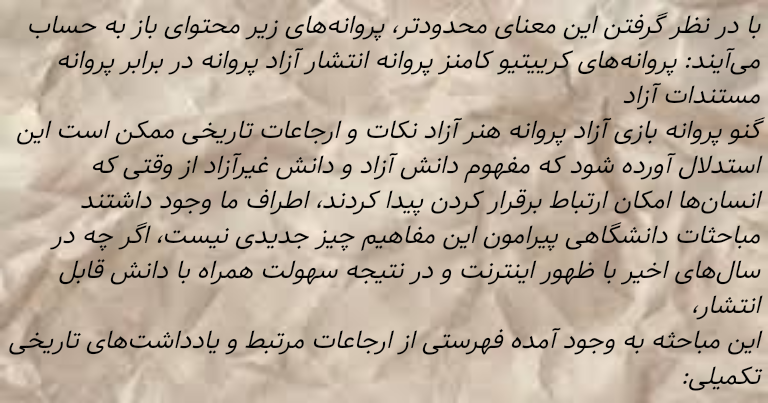}\caption{Page Sample 7}\end{subfigure}\hfill
\begin{subfigure}[b]{0.30\textwidth}\centering\includegraphics[width=\textwidth]{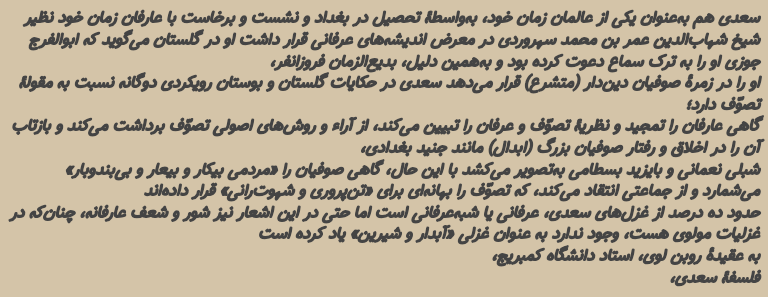}\caption{Page Sample 8}\end{subfigure}\hfill
\begin{subfigure}[b]{0.30\textwidth}\centering\includegraphics[width=\textwidth]{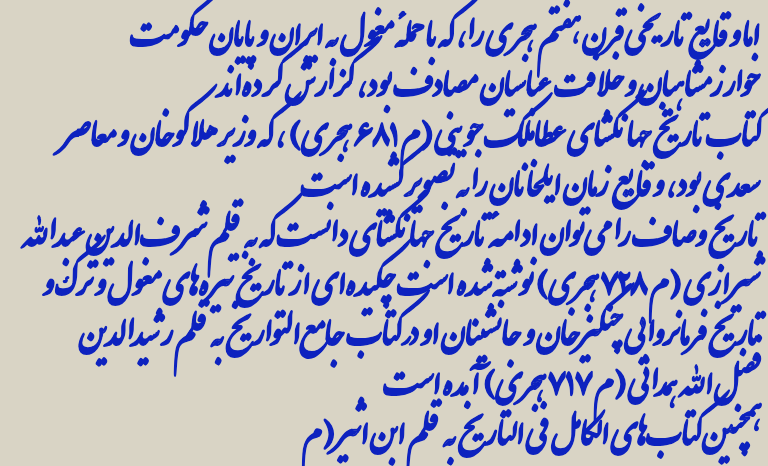}\caption{Page Sample 9}\end{subfigure}
\caption{Representative page-level samples from Persian Pixel, showcasing full-page documents with complex layouts, diverse fonts, and advanced visual degradations.}
\label{fig:page_samples}
\end{figure*}

This appendix offers a qualitative view of the dataset to complement the statistics of Section~\ref{sec:dataset_analysis}. The grids below present every representative sample we release at each granularity: ten at the sentence level (Fig.~\ref{fig:sentence_samples}), eleven at the paragraph level (Fig.~\ref{fig:paragraph_samples}), and nine at the page level (Fig.~\ref{fig:page_samples}), chosen to convey the range of fonts, textual complexity, and stochastic degradation produced by the generation pipeline. Collectively they illustrate the span from pristine digital renders to heavily degraded, historically styled documents that motivates the augmentation design of Section~\ref{sec:augmentation}.

\end{document}